\documentclass[10pt,twocolumn,letterpaper]{article}

\usepackage{cvpr}              

\usepackage{graphicx}
\usepackage{amsmath}
\usepackage{amssymb}
\usepackage{booktabs}
\usepackage{wasysym}
\usepackage{makecell}
\usepackage{multirow}
\usepackage{colortbl}
\usepackage[accsupp]{axessibility}  

\usepackage[pagebackref,breaklinks,colorlinks]{hyperref}

\usepackage{pifont}
\newcommand{\cmark}{\ding{51}}%
\usepackage[dvipsnames]{xcolor}

\usepackage{algorithm}
\usepackage{listings}

\definecolor{linkcolor}{HTML}{ED1C24}
\definecolor{citecolor}{HTML}{0071bc}  
\definecolor{highlightgreen}{HTML}{39b54a}  
\definecolor{highlightred}{HTML}{ea4335}  
\newcommand{\hlg}[1]{\textcolor{highlightgreen}{#1}}

\usepackage[capitalize]{cleveref}
\crefname{section}{Sec.}{Secs.}
\Crefname{section}{Section}{Sections}
\Crefname{table}{Table}{Tables}
\crefname{table}{Tab.}{Tabs.}

\def\cvprPaperID{303} 
\def\confName{CVPR}
\def\confYear{2023}

\definecolor{_yellow}{RGB}{255,217,50}
\definecolor{_darkblue}{RGB}{1,25,147}
\definecolor{_red}{RGB}{238,34,12}
\definecolor{_pink}{RGB}{255,64,255}
\definecolor{_orange}{RGB}{255,147,0}
\definecolor{_blue}{RGB}{4,51,255}
\definecolor{_green}{RGB}{97,216,54}
\definecolor{_purple}{RGB}{122,129,255}
\definecolor{_skyblue}{RGB}{86,193,255}
\definecolor{_coral}{RGB}{255,100,78}
\definecolor{_violet}{RGB}{148,55,250}

\begin{document}

\title{DETR with Additional Global Aggregation \\ for Cross-domain Weakly Supervised Object Detection}

\author{Zongheng Tang$^{1, 2, 3}$ \quad Yifan Sun$^{2}$ \quad Si Liu$^{1,3}$\thanks{Corresponding author: Si Liu} \quad Yi Yang$^{4}$\\
$^1$Institute of Artificial Intelligence, Beihang University \quad $^2$Baidu Inc \\ 
\quad $^3$Hangzhou Innovation Institute, Beihang University \quad $^4$CCAI, Zhejiang University \\
{\tt\small tzhhhh123@buaa.edu.cn \hfill sunyf15@tsinghua.org.cn \hfill liusi@buaa.edu.cn \hfill yangyics@zju.edu.cn}
}

\maketitle

\begin{abstract}
This paper presents a DETR-based method for cross-domain weakly supervised object detection (CDWSOD), aiming at adapting the detector from source to target domain through weak supervision. We think DETR has strong potential for CDWSOD due to an insight: the encoder and the decoder in DETR are both based on the attention mechanism and are thus capable of aggregating semantics across the entire image. The aggregation results, \emph{i.e.}, image-level predictions, can naturally exploit the weak supervision for domain alignment. Such motivated, we propose DETR with additional Global Aggregation (DETR-GA), a CDWSOD detector that simultaneously makes ``instance-level + image-level'' predictions and utilizes ``strong + weak`` supervisions. The key point of DETR-GA is very simple: for the encoder / decoder, we respectively add multiple class queries / a foreground query to aggregate the semantics into image-level predictions. Our query-based aggregation has two advantages. First, in the encoder, the weakly-supervised class queries are capable of roughly locating the corresponding positions and excluding the distraction from non-relevant regions. Second, through our design, the object queries and the foreground query in the decoder share consensus on the class semantics, therefore making the strong and weak supervision mutually benefit each other for domain alignment. Extensive experiments on four popular cross-domain benchmarks show that DETR-GA significantly improves cross-domain detection accuracy (\emph{e.g.}, 29.0\% $\rightarrow$ 79.4\% mAP on PASCAL VOC $\rightarrow$ Clipart$_{\texttt{all}}$ dataset) and advances the states of the art. 
\end{abstract}

\section{Introduction}

\begin{figure}[ht!]
\centering
\vspace{-.5em}
\includegraphics[width=\linewidth]{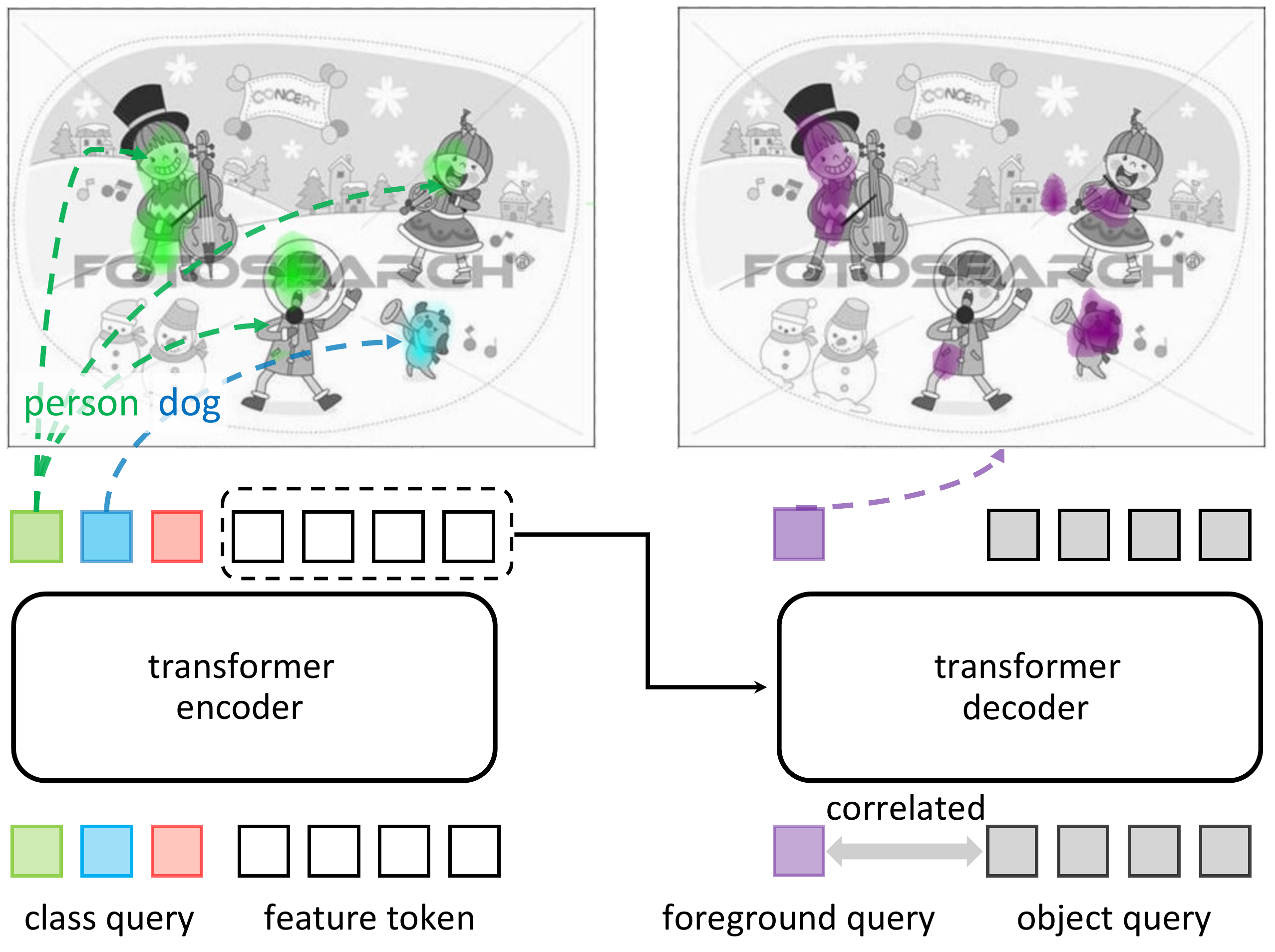}
\vspace{-2em}
\caption{To exploit the weak supervision, DETR-GA aggregates the semantic information across the entire image into image-level predictions. Specifically, DETR-GA adds multiple class queries / a foreground query into the transformer encoder / decoder, respectively. The foreground query is correlated with the object queries but has no position embedding. We visualize the attention score of some queries, \emph{e.g.}, \textcolor{green}{``person``} and \textcolor{blue}{``dog''}.  Despite no position supervision, each class query / the foreground query attends to the class-specific / foreground regions for semantic aggregation.}

\vspace{-1em}
\label{fig:intro}
\end{figure}

The cross-domain problem is a critical challenge for object detection in real-world applications. Concretely, there is usually a domain gap between the training and testing data. This domain gap significantly compromises the detection accuracy when the detector trained on the source domain is directly deployed on a novel target domain. To mitigate the domain gap, existing domain adaptation methods can be categorized into supervised, unsupervised \cite{chen2018domain,saito2019strong,deng2021unbiased}, and weakly supervised approaches \cite{inoue2018cross,hou2021informative,ouyang2021pseudo,xu2022h2fa}. Among the three approaches, we are particularly interested in the weakly supervised one because it requires only image-level annotations and achieves a good trade-off between the adaptation effect and the annotation cost. Therefore, this paper challenges the cross-domain weakly supervised object detection (CDWSOD), aiming at adapting the detector from the source to target domain through weak supervision.

We think the DETR-style detector \cite{zhu2020deformable,carion2020end,li2022dn} has high potential for solving CDWSOD. In contrast to current CDWSOD methods dominated by pure convolutional neural network detectors (``CNN detectors''), this paper is the first to explore DETR-style detectors for CDWSOD, to the best of our knowledge. 
Our optimism for DETR is NOT due to its prevalence or competitive results in generic object detection. In fact, we empirically find the DETR-style detector barely achieves any superiority against CNN detectors for direct cross-domain deployment (Section~\ref{sec:ablation}). Instead, our motivation is based on the insight, \emph{i.e.}, the DETR-style detector has superiority for combining the strong and weak supervision, which is critical for CDWSOD \cite{inoue2018cross,hou2021informative,xu2022h2fa}.

Generally, CDWSOD requires using weak (\emph{i.e.}, image-level) supervision on target domain to transfer the knowledge from source domain. 
Therefore, it is essential to supplement the detector with image-level prediction capability.
We argue that this essential can be well accommodated by two basic components in DETR, \emph{i.e.}, the encoder and the decoder. 
Both the encoder and the decoder are based on the attention mechanism and thus have strong capability to capture long-range dependencies. 
This long-range modeling capability, by its nature, is favorable for aggregating semantic information to make image-level predictions.

To fully exploit the weak supervision in CDWSOD, this paper proposes DETR with additional Global Aggregation (DETR-GA). 
DETR-GA adds attention-based global aggregation into DETR so as to make image-level predictions, while simultaneously preserving the original instance-level predictions. Basically, DETR uses multiple object queries in the decoder to probe local regions and gives instance-level predictions. Based on DETR, DETR-GA makes two simple and important changes: for the encoder / decoder, it respectively adds multiple class queries / a foreground query to aggregate semantic information across the entire image. 
The details are explained below:

\emph{1) The encoder makes image-level prediction through a novel class query mechanism}. Specifically, the encoder adds multiple class queries into its input layer, with each query responsible for an individual class.  
Each class query probes the entire image to aggregate class-specific information and predicts whether the corresponding class exists in the image. During training, we use the image-level multi-class label to supervise the class query predictions. 

Despite NO position supervision, we show these class queries are capable to roughly locate the corresponding position (Fig.~\ref{fig:intro}) and thus exclude the distraction from non-relevant regions. Therefore, our class query mechanism achieves better image-level aggregation effect than the average pooling strategy that is commonly adopted in pure-CNN CDWSOD methods. Empirically, we find this simple component alone brings significant improvement for CDWSOD, \emph{e.g.}, +20.8 mAP on PASCAL VOC $\rightarrow$ Clipart$_{\texttt{test}}$. 

\emph{2) The decoder gives image-level and instance-level predictions simultaneously through correlated object and foreground queries}. To this end, we simply remove the position embedding from an object query and use the remained content embedding as the foreground query. The insight for this design is: 
in a object query, the position embedding encourages focus on local region \cite{li2022dn,meng2021conditional,liu2022dab}, while the content embedding is prone to global responses to all potential foreground regions. Therefore, when we remove the position embedding, an object query discards the position bias and becomes a foreground query with global responses (as visualized in Fig.~\ref{fig:intro}). Except this difference (\emph{i.e.}, with or without position embedding), the object queries and the foreground query share all the other elements in the decoder, \emph{e.g.}, the self-attention layer, cross-attention layer. Such correlation encourages them to share consensus on the class semantic and thus benefits the domain alignment along all the classes.

Overall, DETR-GA utilizes the weak supervision on the encoder and decoder to transfer the detection capability from source to target domain.
Experimental results show that DETR-GA improves cross-domain detection accuracy by a large margin. 
Our main contributions can be summarized as follows:

    $\bullet$ As the first work to explore DETR-style detector for CDWSOD, this paper reveals that DETR has strong potential for weakly-supervised domain adaptation because its attention mechanism can fully exploit image-level supervision by aggregating semantics across the entire image. 
    
    $\bullet$ We propose DETR-GA, a CDWSOD detector that simultaneously makes ``instance-level + image-level'' predictions and can utilize both ``strong + weak'' supervision. The key point of DETR-GA is the newly-added class queries / foreground query in the encoder / decoder, which promotes global aggregation for image-level prediction.

    $\bullet$ Extensive experiments on four popular cross-domain benchmarks show that DETR-GA significantly improves CDWSOD accuracy and advances the states of the art. For example, on PASCAL VOC $\rightarrow$  Clipart$_{\texttt{all}}$, DETR-GA improves the baseline from 29.0\% to 79.4\% . 
\begin{figure*}[ht]
\centering
\vspace{-0.8em}
\includegraphics[width=\linewidth]{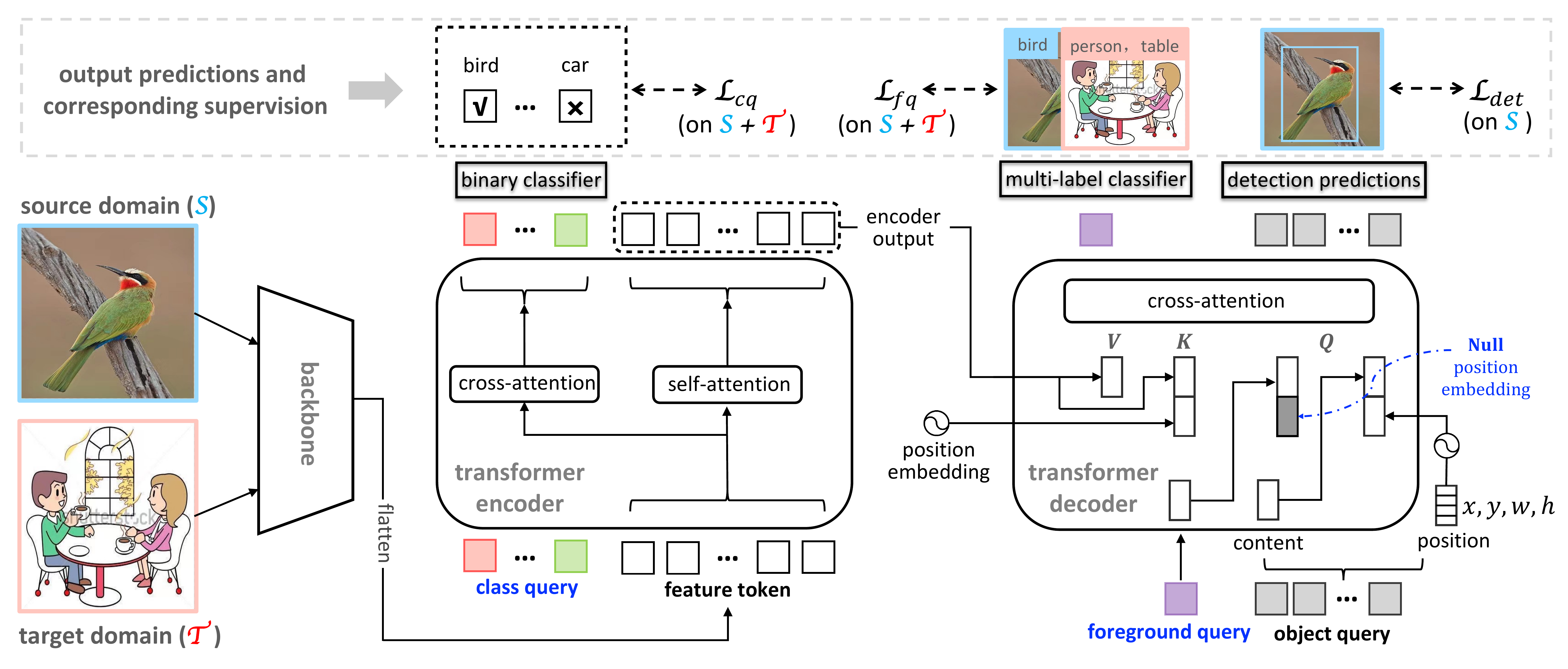}
\vspace{-2em}
\caption{Overview of the proposed DETR-GA. DETR-GA 
tries to transfer the detection knowledge from the fully-labeled \textcolor{black}{source} domain \textcolor{cyan}{$\mathcal{S}$} to the \textcolor{black}{target} domain \textcolor{red}{$\mathcal{T}$} through weak supervision. Based on a DETR-style detector consisting of backbone, encoder and decoder, DETR-GA adds two modules, \emph{i.e.}, \textcolor{blue}{class query} and \textcolor{blue}{foreground query}.
In the encoder, multiple class queries aggregate class-specific information from all the feature tokens through cross-attention. In the decoder, the foreground query and the object queries are correlated with each other, because they share most processing except that the foreground query has \textcolor{blue}{null} position embedding. Correspondingly, they make image-level and instance-level predictions, respectively. 
Overall, we utilize weak supervision / hybrid (weak + strong) supervision for the encoder / decoder, therefore aligning the class semantics between the source and target domain.
}
\vspace{-1em}
\label{fig:pipeline}
\end{figure*}

\section{Related Work}
\paragraph{Object detection.}
Object detection is a fundamental task in computer vision and attract much attention for years.
The pure CNN-based detectors can be roughly divided into two-stage methods, \emph{e.g.}, Faster RCNN~\cite{ren2015faster}, and single-stage methods,\emph{e.g.}, YOLO~\cite{redmon2016you} and FCOS~\cite{tian2019fcos}.
Although these CNN-based detectors have made significant advancements, they still require numerous hand-designed components, \emph{e.g.}, non-maximum suppression (NMS), and anchor generation. 
Recently, DETR~\cite{carion2020end} propose an end-to-end object detection pipeline without the above hand-designed components.
Many DETR-style detectors~\cite{li2022dn,zhu2020deformable,meng2021conditional,liu2022dab} are then proposed to further improve DETR. 
For example, DAB-DETR~\cite{liu2022dab} proposes to replace the original position embedding with 4d box anchors.
DN-DETR\cite{li2022dn} proposes a denoising training method to accelerate convergence.

So far as we know, this paper is the first to explore DETR-style detector for CDWSOD. During our preliminary experiments, we find that DETR presents no obvious superiority for direct cross domain deployment, though it usually surpasses the pure-CNN detectors on generic object detection. For example, on PASCAL VOC $\rightarrow$ Clipart$_{\texttt{all}}$, Faster RCNN and DETR respectively achieve 29.0. and 29.2, which are comparable. However, we still believe that DETR has strong potential for CDWSOD, because its attention mechanism promotes global aggregation and can better exploit weak supervision. Empirically, we show that the proposed DETR-GA achieves significant improvement over direct cross-domain deployment, and surpasses the strongest pure-CNN method by a clear margin(though they achieve similar direct cross-domain deployment accuracy).

\paragraph{Global aggregation in CDWSOD.}
Aggregating the semantic information into image-level prediction is critical for cross-domain alignment in CDWSOD~\cite{xu2022h2fa,hou2021informative}.
Xu~\cite{xu2022h2fa} uses average pooling for global aggregation on the backbone feature and proposes a softmax-based aggregation to merge all the instance-level predictions from the detection head.
ICCM~\cite{hou2021informative} divides each image into semantic clusters and aligns the foreground regions in the target domain with the labeled region in the source domain.
In contrast, in DETR-GA, the class and foreground queries utilize attention mechanisms for global aggregation. The query-based global aggregation has an advantage, \emph{i.e.}, the query is capable of roughly locating the corresponding regions despite no position supervision.

\paragraph{Cross-domain object detection.}
Cross-domain object detection (CDOD) aims to adapt the detector from the fully-labeled source domain to the label-scarce target domain. 
Previous works mainly focus on the unsupervised domain adaptation object detection(UDAOD) and related fields~\cite{su2020adapting,zhao2020collaborative,wang2021domain,shen2021cdtd,liu2021domain,zhang2021densely,hu2023suppressing,tao2022exploring,tang2021human,he2022cross,he2019multi,he2020domain,cai2019exploring,zhu2019adapting,gu2021pit,deng2021unbiased,vs2021mega,chen2021dual,}.
The UDAOD methods can be roughly categorized into image translation~\cite{HanKaiHsu2019ProgressiveDA,TaekyungKim2019DiversifyAM}, self-supervision~\cite{roychowdhury2019automatic,khodabandeh2019robust,kim2019self,ramamonjison2021simrod,li2021category}, and adversarial training~\cite{chen2018domain,chen2021scale,saito2019strong,zhao2020adaptive,hsu2020every,zheng2020cross,xu2020cross,wang2021exploring,wu2021vdd,zhang2021rpn,tian2021knowledge,rezaeianaran2021seeking,chen2021i3net}.
Recently, SFA~\cite{wang2021exploring} utilize adversarial training for domain alignment based on the DETR-style detector.
Compared with UDAOD, CDWSOD~\cite{inoue2018cross,hou2021informative,ouyang2021pseudo} offers more image-level annotations on the target domain.
DETR-GA utilizes the additional global aggregation to make image-level domain alignment and achieves better performance.

\section{DETR-GA}
\subsection{Overview}
\textbf{Task definition}. Unlike weakly supervised object detection\cite{yang2019towards,tang2017multiple,bilen2016weakly,wan2019min} only utilize the weak supervision on the target domain,
CDWSOD aims at adapting the detector from the source domain to the target domain through weak supervision. In the source domain, each image has instance-level annotations with object classes and bounding boxes, while in the target domain, each image has only multi-class labels indicating which classes exist. 

\textbf{DETR Revisit}. Since DETR-GA is based on the DETR pipeline, we first give a brief revisit to DETR. 
DETR consists of a backbone model, a transformer encoder and a transformer decoder. Given an input image, DETR first uses a backbone model (\emph{e.g.}, ResNet-50 \cite{he2016deep}) to extract the feature map. Afterward, DETR flattens the feature map into feature tokens and feeds them into the transformer encoder. 
In the encoder, all the feature tokens are updated through self-attention block by block. The final output states of these feature tokens flow into the transformer decoder, along with multiple (\emph{i.e.}, 300) learnable object queries. 
In each decoder block, object queries first perform self-attention between themselves and then cross-attention to the feature tokens. For simplicity, Fig.~\ref{fig:pipeline} only presents the cross-attention part. Please refer to the supplementary for a complete view of the DETR decoder. After multiple blocks of self-attention and cross-attention, the output states of the object queries generate instance-level detection predictions (\emph{i.e.}, the object class end bounding box position), which are supervised through a detection loss $\mathcal{L}_{det}$\cite{carion2020end}.

\textbf{Overall structure of DETR-GA}. 
As shown in Figure~\ref{fig:pipeline}, DETR-GA takes the mixture of source and target images as its input. 
To fully exploit the weak supervision, DETR-GA supplements the original detector with image-level predictions in both the encoder and decoder. 
Given the image-level prediction capability of the encoder and decoder, DETR-GA facilitates joint training on the source and target domain. 
The encoder aligns the class semantics across the source and target domain through weak supervision. Afterward, the decoder further delivers the object detection capability from source to target through hybrid supervision (\emph{i.e.}, strong supervision on the source and weak supervision on the target). 

Specifically, for the encoder, we add multiple class queries into its input layer, with each query responsible for an individual class. 
Each class query searches the entire image and aggregates class-specific information through cross-attention. The aggregated outputs predict whether the corresponding classes exist in the image and are supervised with a binary cross-entropy loss $\mathcal{L}_{cq}$.
For the decoder, we add a foreground query into its input layer. The foreground shares the content embedding with all the object queries and has NO position embedding. Therefore, it can attend to all the potential foreground without positional bias. Its aggregation output predicts which classes exist in the image and uses another multi-class binary cross-entropy loss $\mathcal{L}_{fq}$ for supervision. 
In the encoder, the aggregated semantic information is naturally used for image-level prediction and can be trained through weak supervision. Importantly, in the decoder, the foreground query is correlated with the object queries, making the image-level and instance-level predictions mutually benefit each other.

Given the additional image-level predictions from the encoder and decoder, DETR-GA is capable to combine weak and strong supervision. Specifically, the encoder receives weak supervision from ``source+target'' domains ($\mathcal{L}_{cq}$). The decoder receives weak supervision from ``source+target'' domains ($\mathcal{L}_{fq}$), and meanwhile receives strong supervision from the target domain ($\mathcal{L}_{det}$). Moreover, we further add some common practices ~\cite{xu2022h2fa,wang2021exploring} on the backbone, \emph{i.e.}, using an adversarial loss $\mathcal{L}_{dc}$ for domain-level alignment and an image-level classification loss $\mathcal{L}_{bc}$. 

During training, we combine all the described loss functions to optimize DETR-GA in an end-to-end manner: 

\begin{equation}
\small
\mathcal{L}= 
\underbrace{\vphantom{\lambda_\textit{cq}\mathcal{L}_\textit{cq}}\lambda_\textit{bc}\mathcal{L}_\textit{bc}+\lambda_\textit{dc}\mathcal{L}_\textit{dc}}_\text{backbone}+
\underbrace{\lambda_\textit{cq}\mathcal{L}_\textit{cq}}_\text{encoder}+
\underbrace{\lambda_\textit{fq}\mathcal{L}_\textit{fq}+
\mathcal{L}_\textit{det}}_\text{decoder}.
\label{eq:loss_all}
\end{equation}

During inference, DETR-GA does not need the image-level predictions and thus resumes the standard DETR ~\cite{carion2020end} pipeline without extra computational cost.

\subsection{Encoder with Class Queries}
We add $C$ (the number of total classes) class queries $\mathbf{Q}=\{\mathbf{q}_1, \mathbf{q}_2, \cdots, \mathbf{q}_C\}$ into the encoder. Each class query $\mathbf{q}_i$ is a learnable vector (\emph{i.e.}, $\mathbf{q}_i \in{\mathbb{R}}^{d}$) and corresponds to the $i$-th class. We will use the output state of $\mathbf{q}_i$ as a binary indicator to identify the presence of the $i$-th class. 

Specifically, let us denote the collection of all the feature tokens as $\mathbf{T}$, and their position embedding as $\mathbf{P}_T$. In each transformer block of the encoder,  all the class queries $\mathbf{Q}$ are updated through cross-attention to the feature tokens $\mathbf{T}$, in parallel to the self-attention update of the feature tokens. Formally, in the $k$-th ($k=1,2, \cdots, K$) transformer block of the encoder, the self-attention and the cross-attention update are formulated as:

\begin{footnotesize}
\begin{equation}\label{eqn:encoder}
\begin{split}
 \mathbf{T}^k &= \mathbf{T}^{k-1} + attn(\mathbf{T}^{k-1} + \mathbf{P}_T, \mathbf{T}^{k-1} + \mathbf{P}_T, \mathbf{T}^{k-1}) \\
 \mathbf{Q}^k &= \mathbf{Q}^{k-1} + attn(\mathbf{Q}^{k-1}, \mathbf{T}^{k-1} + \mathbf{P}_T, \mathbf{T}^{k-1}), 
\end{split}
\end{equation}
\end{footnotesize}
where $attn(\texttt{query}, \texttt{key}, \texttt{value})$ denotes the attention layer \cite{vaswani2017attention}, the superscript $k$ indicates the $k$-th transformer block. There are a layer norm and a feed-forward network (FFN) \cite{vaswani2017attention} following the attention layer and are omitted here for brevity.

Given the output states of the class query, \emph{i.e.}, $\mathbf{Q}^K=\{\mathbf{q}^K_1, \mathbf{q}^K_2, \cdots, \mathbf{q}^K_C\}$ ($K$ is the transformer depth), the encoder uses a binary classifier to predict the presence of the $i$-th class by:

\begin{equation} \label{eqn:encoder_predict}
    p_i = \texttt{sigmoid}\{\mathbf{q}^K_i \cdot \mathbf{w}_i\},
\end{equation}
where $\mathbf{w}_i$ is the weight vector for the $i$-th class in the classification head, `` $\cdot$  '' is the inner product operation. We note that Eqn.~\ref{eqn:encoder_predict} has a minor difference against the canonical multi-class predictor, which uses a single input to simultaneously predict the presence of multiple classes. In contrast,  Eqn.~\ref{eqn:encoder_predict} has multiple inputs, \emph{i.e.}, each $\mathbf{q}^K_i$ is responsible for an individual class.

The above classifier is supervised by a standard binary cross-entropy (BCE) loss 
$\mathcal{L}_\textit{cq}$, which is formulated as:
\begin{equation} \label{eqn:encoder_loss}
\begin{split}
\mathcal{L}_\textit{cq} = \frac{1}{C}\sum_{i=1}^C [y_ilogp_i+(1-y_i)(1-logp_i)],
\end{split}
\end{equation}
where $y_i$ is set to 1 if class $c_i$ exists, otherwise 0. 

We provide an intuitive review of the domain alignment procedure in the encoder. 
Through the cross-attention in Eqn.~\ref{eqn:encoder}, each class query absorbs information from all the feature tokens, therefore facilitating global aggregation. The aggregation results $\mathbf{Q}^M$ are used to predict the presence of each class through Eqn.~\ref{eqn:encoder_predict}. The  corresponding weak supervision (Eqn.~\ref{eqn:encoder_loss}) is enforced on both the source and target domain and thus aligns each class across two domains. 

An important advantage of our query-based aggregation is: though there is no position supervision, each class query is capable of roughly locating its object regions, therefore excluding the distraction from non-relevant regions, as shown in Figure~\ref{fig:intro} and in the supplementary. In contrast, CNN CDWSOD methods usually use global average pooling \cite{xu2022h2fa} for aggregation. It inevitably introduces interference among different classes and thus compromises the domain alignment effect. In an ablation study (Section~\ref{sec:ablation}), we show that our query-based aggregation is significantly better (\emph{e.g.}, 9.4 mAP higher on PASCAL VOC $\rightarrow$ Clipart$_{\texttt{test}}$) than the average pooling aggregation.

\subsection{Decoder with Foreground and Object Queries}
In DETR-GA, the transformer decoder has three types of input, \emph{i.e.}, feature tokens ($\mathbf{T}^K$ output from the encoder), a newly-added foreground query $\mathbf{f}\in{\mathbb{R}}^{d}$ (which is a vector), and $N$ object queries. 
The object queries consist of the content embedding $\mathbf{O}=\{\mathbf{o}_1, \mathbf{o}_2, \cdots, \mathbf{o}_N\}$ and the  position embedding  $\mathbf{P}_O\in{\mathbb{R}}^{N \times d}$.

In the $l$-th ($l-1,2, \cdots, L$ ) transformer block of the decoder, the foreground query and the object queries simultaneously undergo a self-attention update and a following cross-attention update. 

The self-attention update is formulated as:
\begin{equation}\label{eqn:decoder_self}
\begin{split}
[\widetilde{\mathbf{O}}^{l-1}, \widetilde{\mathbf{f}}^{l-1}]= self([\mathbf{O}^{l-1} + \mathbf{P}_O^{l-1}, f^{l-1}]),
\end{split}
\end{equation}
where ``$[\quad]$'' is the concatenation operation, $self$ is the self-attention layer. We adopt the same self-attention layer as in DN-DETR \cite{li2022dn} and provide a detailed description in the supplementary.

The following cross-attention update is formulated as: 

\begin{footnotesize}
\begin{equation}\label{eqn:decoder_across}
\begin{split}
 \mathbf{O}^l &= \mathbf{O}^{l-1} +  attn(H(\widetilde{\mathbf{O}}^{l-1}, \mathbf{P}_O^{l-1}), H(\mathbf{T}^M, \mathbf{P}_T), \mathbf{T}^M),  \\
 \mathbf{f}^l 
 &= \mathbf{f}^{l-1} + attn(H(\widetilde{\mathbf{f}}^{l-1}, \mathbf{0}), H(\mathbf{T}^M, \mathbf{P}_T), \mathbf{T}^M),
\end{split}
\end{equation}
\end{footnotesize}
where $attn(\texttt{query}, \texttt{key}, \texttt{value})$ denotes the attention layer, $H$ is the operation for entangling the content and position embedding (\emph{e.g.}, the concatenation operation as in DAB-DETR \cite{liu2022dab} and DN-DETR \cite{li2022dn}).
The foreground query has a null position embedding, as illustrated in Fig.~\ref{fig:pipeline} and thus may be viewed as removing the impact of position embedding.
The position embedding update is the same as in DN-DETR and is omitted here.

The above self-attention (Eqn.~\ref{eqn:decoder_self}) and cross-attention (Eqn.~\ref{eqn:decoder_across}) are almost the same as in DAB-DETR and DN-DETR), except for adding a foreground query with no position embedding. Although we adopt the DN-DETR as the direct baseline, we note that Eqn.~\ref{eqn:decoder_self} and Eqn.~\ref{eqn:decoder_across} jointly depict a general DETR decoder and are compatible with other DETR-style detectors. \textcolor{black}{For example, $H$ for entangling the content embedding and position embedding can be modified to ``$+$'' operation, as in the original DETR.} 

Albeit simple, there are two good characteristics, \emph{i.e.}, removing the position embedding for global attention and correlating the foreground and object queries:

$\bullet$ The foreground query has no position embedding and thus favors global aggregation. In DETR-style detectors, the position embedding encourages the object query to spatially focus on the local region. In fact, a keynote of recent state-of-the-art DETR progress \cite{liu2022dab,li2022dn} is to improve the position embedding for localizing the objects.  Removing the position embedding removes the potential position bias and encourages $\mathbf{f}$ to consider only the semantic content for information aggregation. Empirically, we find that adding position embedding to the foreground query compromises DETR-GA (Section~\ref{sec:ablation}).

$\bullet$ The foreground query is strongly correlated with the object queries because they share almost all the update elements (\emph{e.g.}, the self-attention layer, the cross-attention layer, and the same feature tokens). Intuitively, in Eqn.~\ref{eqn:decoder_self} and Eqn.~\ref{eqn:decoder_across}, the foreground query may be viewed as a specific object query with ZERO position embedding. This correlation encourages the foreground query and object queries to share consensus on the class semantics and thus benefits class-wise alignment. Empirically, we find removing this correlation compromises DETR-GA (Section~\ref{sec:ablation}).

Consequently, the encoder simultaneously gives multiple instance-level predictions (from the object query output) and an image-level prediction (from the foreground query output). 
These predictions are respectively supervised through $\mathcal{L}_{det}$ and $\mathcal{L}_{fq}$, as shown in Fig.~\ref{fig:pipeline}. $\mathcal{L}_{fq}$ is a standard BCE loss as in Eqn.~\ref{eqn:encoder_loss}. 
\begin{table*}[t!]
\renewcommand\arraystretch{1.1}
\centering
\vspace{-0.5em}
\scriptsize
\setlength{\tabcolsep}{3.8pt}
\setlength{\aboverulesep}{0pt} 
\setlength{\belowrulesep}{0pt} 
\begin{tabular}{l|cccccccccccccccccccc|c}
\toprule
\hspace{-0.4em}Method & aero & bike & bird & boat & bottle & bus & car & cat & chair & cow & table & dog & horse & mbike & person & plant & sheep & sofa & train & tv & mean \\
\Xhline{2\arrayrulewidth}
\hspace{-0.4em}\textit{\textcolor{gray}{WS Group}} & & & & & & & & & & & & & & & & & & & & \\
PCL~\cite{8493315} & 3.4 & 10.6 & 2.3 & 1.7 & 5.2 & 3.4 & 23.3 & 1.2 & 5.6 & 0.4 & 7.8 & 3.7 & 5.6 & 0.3 & 24.5 & 19.7 & 11.9 & 3.6 & 9.2 & 25.4 & 8.4 \\
EDRN~\cite{DRN-WSOD_2020_ECCV} & 2.7 & 13.5 & 1.2 & 4.2 & 1.8 & 10.3 & 25.7 & 0.4 & 8.4 & 0.3 & 3.2 & 2.7 & 1.1 & 0.7 & 29.4 & 17.2 & 5.2 & 1.6 & 2.9 & 19.1 & 7.6 \\
\hline
\hspace{-0.4em}\textit{\textcolor{gray}{UDA Group}} & & & & & & & & & & & & & & & & & & & & \\
SWDA~\cite{saito2019strong} & 26.2 & 48.5 & 32.6 & 33.7 & 38.5 & 54.3 & 37.1 & 18.6 & 34.8 & 58.3 & 17.0 & 12.5 & 33.8 & 65.5 & 61.6 & 52.0 & 9.3 & 24.9 & 54.1 & 49.1 & 38.1 \\
HTD~\cite{chen2020harmonizing} & 33.6 & 58.9 & 34.0 & 23.4 & 45.6 & 57.0 & 39.8 & 12.0 & 39.7 & 51.3 & 21.1 & 20.1 & 39.1 & 72.8 & 63.0 & 43.1 & 19.3 & 30.1 & 50.2 &  51.8 & 40.3 \\
IIOD~\cite{9362301} & 41.5 & 52.7 & 34.5 & 28.1 & 43.7 & 58.5 & 41.8 & 15.3 & 40.1 & 54.4 & 26.7 & 28.5 & 37.7 & 75.4 & 63.7 & 48.7 & 16.5 & 30.8 & 54.5 & 48.7 & 42.1 \\
I$^{3}$Net~\cite{chen2021i3net} & 30.0 & 67.0 & 32.5 & 21.8 & 29.2 & 62.5 & 41.3 & 11.6 & 37.1 & 39.4 & 27.4 & 19.3 & 25.0 & 67.4 & 55.2 & 42.9 & 19.5 & 36.2 & 50.7 & 39.3 & 37.8 \\
DBGL~\cite{chen2021dual} & 28.5 & 52.3 & 34.3 & 32.8 & 38.6 & 66.4 & 38.2 & 25.3 & 39.9 & 47.4 & 23.9 & 17.9 & 38.9 & 78.3 & 61.2 & 51.7 & 26.2 & 28.9 & 56.8 & 44.5 & 41.6 \\
TIA~\cite{zhao2022task} & 42.2 & 66.0 & 36.9 & 37.3 & 43.7 & 71.8 & 49.7 & 18.2 & 44.9 & 58.9 & 18.2 & 29.1 & 40.7 & 87.8 & 67.4 & 49.7 & 27.4 & 27.8 & 57.1 & 50.6 & 46.3 \\
\hline
\hspace{-0.4em}\textit{\textcolor{gray}{CDWS Group}} & & & & & & & & & & & & & & & & & & & & \\
DT+PL~\cite{inoue2018cross} & 50.1 & 75.0 & 37.0 & 38.7 & 58.1 & 83.4 & 50.1 & 38.0 & 55.2 & 67.3 & 51.1 & 34.8 & 49.8 & 89.9 & 60.2 & 63.4 & 28.8 & 42.4 & 62.6 & 70.9 & 55.3 \\
ICCM~\cite{hou2021informative} & 39.8 & 66.7 & 37.2 & 42.5 & 43.3 & 48.1 & 48.1 & 21.3 & 46.5 & 73.0 & 29.0 & 29.8 & 57.3 & 78.6 & 67.8 & 48.7 & 46.3 & 19.3 & 42.8 & 48.5 & 46.7 \\
H$^{2}$FA~\cite{xu2022h2fa}  & 58.1 & 73.0 & 56.8 & 50.4 & 61.2 & 98.6 & 69.5 & 57.8 & 66.4 & 77.1 & 56.1 & 84.1 & 64.3 & 100.0 & 78.1 & \textbf{78.2} & 43.5 & 65.4 & 77.3 & 79.7 & 69.8 \\

\rowcolor[RGB]{220,220,220}DETR-GA  & \textbf{86.7} & \textbf{85.6}  & \textbf{64.2} & \textbf{66.4} & \textbf{67.4} & \textbf{100.0} & \textbf{75.5} & \textbf{70.0} & \textbf{72.2} & \textbf{81.4} & \textbf{83.8} & \textbf{87.0} & \textbf{75.2} & \textbf{100.0} & \textbf{82.8} & 75.5 & \textbf{67.8} & \textbf{76.1} & \textbf{86.2} & \textbf{85.1} & \textbf{79.4} \\
\bottomrule
\end{tabular}
\vspace{-1em}
\caption{Mean AP performance (\%) on Clipart$_{\texttt{all}}$.}
\label{tab:clipart}
\vspace{-.5em}
\end{table*}

\begin{table*}[t!]
\renewcommand\arraystretch{1.1}
\centering
\scriptsize
\setlength{\tabcolsep}{3.8pt}
\setlength{\aboverulesep}{0pt} 
\setlength{\belowrulesep}{0pt} 
\begin{tabular}{l|cccccccccccccccccccc|c}
\toprule
\hspace{-0.4em}Method & aero & bike & bird & boat & bottle & bus & car & cat & chair & cow & table & dog & horse & mbike & person & plant & sheep & sofa & train & tv & mean \\
\Xhline{2\arrayrulewidth}
\hspace{-0.4em}\textit{\textcolor{gray}{WS Group}} & & & & & & & & & & & & & & & & & & & & \\
WSDDN~\cite{bilen2016weakly} & 1.6 & 3.6 & 0.6 & 2.3 & 0.1 & 11.7 & 4.5 & 0.0 & 3.2 & 0.1 & 2.8 & 2.3 & 0.9 & 0.1 & 14.4 & 16.0 & 4.5 & 0.7 & 1.2 & 18.3 & 4.4 \\ 
CLNet~\cite{kantorov2016contextlocnet} & 3.2 & 22.3 & 2.2 & 0.7 & 4.6 & 4.8 & 17.5 & 0.2 & 4.8 & 1.6 & 6.4 & 0.6 & 4.7 & 0.6 & 12.5 & 13.1 & 14.1 & 4.1 & 8.0 & 29.7 & 7.8 \\ 
\hline
\hspace{-0.4em}\textit{\textcolor{gray}{UDA Group}} & & & & & & & & & & & & & & & & & & & & \\
DM~\cite{kim2019diversify} & 28.5 & 63.2 & 24.5 & 42.4 & 47.9 & 43.1 & 37.5 & 9.1 & 47.0 & 46.7 & 26.8 & 24.9 & 48.1 & 78.7 & 63.0 & 45.0 & 21.3 & 36.1 & 52.3 & 53.4 & 41.8 \\
ATF~\cite{he2020domain} & 41.9 & 67.0 & 27.4 & 36.4 & 41.0 & 48.5 & 42.0 & 13.1 & 39.2 & 75.1 & 33.4 & 7.9 & 41.2 & 56.2 & 61.4 & 50.6 & 42.0 & 25.0 & 53.1 & 39.1 &  42.1 \\
UMT~\cite{deng2021unbiased} & 39.6 & 59.1 & 32.4 & 35.0 & 45.1 & 61.9 & 48.4 & 7.5 & 46.0 & 67.6 & 21.4 & 29.5 & 48.2 & 75.9 & 70.5 & 56.7 & 25.9 & 28.9 & 39.4 & 43.6 & 44.1 \\
AT~\cite{li2022cross} & 33.8 & 60.9 & 38.6 & 49.4 & 52.4 & 53.9 & 56.7 & 7.5 & 52.8 & 63.5 & 34.0 & 25.0 & 62.2 & 72.1 & 77.2 & 57.7 & 27.2 & 52.0 & 55.7 & 54.1 & 49.3 \\
\hline
\hspace{-0.4em}\textit{\textcolor{gray}{CDWS Group}} & & & & & & & & & & & & & & & & & & & & \\
DT+PL~\cite{inoue2018cross} & 51.6 & \textbf{84.0} & 30.0 & 41.1 & 52.3 & 82.0 & 50.2 & 19.0 & 51.8 & 58.3 & 41.3 & 14.6 & 47.0 & 86.2 & 61.9 & 58.6 & 24.9 & 22.5 & 47.4 & 52.8 & 48.9 \\
PLGE~\cite{ouyang2021pseudo} & 43.4 & 52.5 & 29.4 & 40.1 & 30.4 & 71.9 & 54.9 & 3.6 & 52.4 & 73.8 & \textbf{53.5} & 24.0 & \textbf{54.8} & \textbf{89.1} & 65.1 & 40.5 & 32.3 & 33.8 & 45.4 & 61.0 & 47.6 \\
H$^{2}$FA\cite{xu2022h2fa} & 38.5 & 70.6 & 38.9 & 47.4 & \textbf{59.6} & \textbf{83.5} & 47.0 & 29.3 & 51.5 & \textbf{76.3} & 44.4 & 48.1 & 47.3 & 79.2 & 75.7 & 54.4 & \textbf{53.9} & 32.0 & 56.6 & 51.1 & 55.3 \\

\rowcolor[RGB]{220,220,220}DETR-GA  & \textbf{64.1} & 63.0 & \textbf{42.7} & \textbf{52.5} & 56.6 & 82.8 & \textbf{57.5} & \textbf{36.3} & \textbf{56.6} & 67.1 & 48.2 & \textbf{53.5} & 41.9 & 84.0 & \textbf{78.0} & \textbf{62.3} & 48.9 & \textbf{48.4} & \textbf{75.2} & \textbf{64.4} & \textbf{59.2} \\

\bottomrule
\end{tabular}
\vspace{-1em}
\caption{Mean AP performance (\%) on Clipart$_{\texttt{test}}$. }
\label{tab:cliparttest}
\vspace{-1.5em}
\end{table*}

\begin{table}[t]
\centering
\scriptsize
\renewcommand\arraystretch{1.1}
\setlength{\tabcolsep}{9pt}
\setlength{\aboverulesep}{0pt} 
\setlength{\belowrulesep}{0pt}
\begin{tabular}{l|c|ccc}
\toprule
\textcolor{gray}{Training Data}& Source Domain & \multicolumn{3}{c}{Target Domain} \\
\hline
\textcolor{gray}{Annotions}   & BBox   & BBox   & Tag    & None  \\
\Xhline{2\arrayrulewidth}
WSOD        &        &        & \cmark &  \\
UDAOD       & \cmark &        &        & \cmark     \\
CDWSOD      & \cmark &        & \cmark & \\
\bottomrule
\end{tabular}
\vspace{-1em}
\caption{The annotations type of different tasks. For example, WSOD methods use the target domain data with image-level tag annotations for training.}
\label{tab:intro_methods}
\vspace{-2em}
\end{table}

\section{Experiments}
\subsection{Dataset}
Following existing works~\cite{saito2019strong,chen2020harmonizing,hou2021informative,inoue2018cross,xu2022h2fa}, we use four datasets, \emph{i.e.}, PASCAL VOC (VOC)~\cite{Everingham10}, Clipart, Watercolor and Comic~\cite{inoue2018cross} for method evaluation. 

The general object detection dataset VOC is used as the source domain with instance-level annotations. Specifically, we use the \texttt{trainval} split of VOC 0712 as the source-domain training data, which provides $\sim$16.5k real-world images of 20 object categories. 

The other three artistic painting datasets are used as the target domains with image-level annotations. Clipart has a \texttt{train} split and a \texttt{test} split, both of which contain 500 images of 20 object categories. There are two popular modes on this dataset: Clipart$_{\texttt{all}}$ merges the two splits for both training and testing~\cite{saito2019strong,chen2020harmonizing,hou2021informative} , while Clipart$_{\texttt{test}}$ uses the former split for training and the latter split for testing~\cite{inoue2018cross,ouyang2021pseudo,deng2021unbiased}.
Both Watercolor and Comic contain 2k images of 6 classes. We use 1k images from the \texttt{train} split for training and 1k images from the \texttt{test} split for testing.

\begin{table}[t!]
\renewcommand\arraystretch{1.1}
\centering
\scriptsize
\setlength{\tabcolsep}{6.3pt}
\setlength{\aboverulesep}{0pt} 
\setlength{\belowrulesep}{0pt} 
\begin{tabular}{l|cccccc|c}
\toprule
\hspace{-0.5em}Method & bike & bird & car & cat & dog & person & mean \\
\Xhline{2\arrayrulewidth}
\hspace{-0.5em}\textit{\textcolor{gray}{WS Group}} & & & & & & & \\
PCL~\cite{8493315} & 1.2 & 0.4 & 8.9 & 2.9 & 2.3 & 15.6 & 5.2 \\ 
EDRN~\cite{DRN-WSOD_2020_ECCV} & 1.6 & 0.5 & 13.2 & 7.2 & 2.5 & 13.2 & 6.4 \\
\hline
\hspace{-0.5em}\textit{\textcolor{gray}{UDA Group}} & & & & & & & \\
SWDA~\cite{saito2019strong} & 30.3 & 19.6 & 28.8 & 15.2 & 24.9 & 46.9 & 27.6 \\
HTD~\cite{chen2020harmonizing} & 35.4 & 14.8 & 26.6 & 13.7 & 26.9 & 40.0 & 26.2 \\
MCAR~\cite{zhao2020adaptive} & 47.9 & 20.5 & 37.4 & 20.6 & 24.5 & 50.2 & 33.5 \\
I$^3$Net~\cite{chen2021i3net} & 47.5 & 19.9 & 33.2 & 11.4 & 19.4 & 49.1 & 30.1 \\
DBGL~\cite{chen2021dual} & 35.6 & 20.3 & 33.9 & 16.4 & 26.6 & 45.3 & 29.7 \\
\hline
\hspace{-0.5em}\textit{\textcolor{gray}{CDWS Group}} & & & & & & & \\
DT+PL~\cite{inoue2018cross} & 53.0 & 23.7 & 34.4 & 27.4 & 27.2 & 44.0 & 35.0 \\
ICCM~\cite{hou2021informative} & 50.6 & 23.3 & 35.4 & 32.3 & 33.8 & 47.1 & 37.1 \\
PLGE~\cite{ouyang2021pseudo} & 55.0 & 21.2 & 40.0 & 35.1 & 37.9 & 60.9 & 41.7 \\
H$^{2}$FA~\cite{xu2022h2fa} & 55.3 & 26.6 & \textbf{45.9} & 38.1 & 45.6 & \textbf{66.8} & 46.4 \\
\rowcolor[RGB]{220,220,220}DETR-GA & \textbf{57.4} & \textbf{36.1} & 39.9 & \textbf{42.0} & \textbf{49.4} & 66.0 & \textbf{48.5} \\
\bottomrule
\end{tabular}
\vspace{-1em}
\caption{Mean AP performance (\%) on Comic.}
\label{tab:comic}
\vspace{-1em}
\end{table}

\subsection{Implementation details}
\textbf{Baseline.} We choose the DETR version of DN-DETR\cite{li2022dn}, which improves the convergence speed and accuracy of the original DETR\cite{carion2020end}, as our baseline. We adopt the evaluation metrics in \texttt{Detectron2}~\cite{wu2019detectron2} for fair comparisons with previous works\cite{xu2022h2fa, inoue2018cross}. We use  ImageNet~\cite{deng2009imagenet} pre-trained ResNet-101~\cite{he2016deep} as the backbone in all experiments unless specified.
The loss weights $\lambda_\textit{cq}$, $\lambda_\textit{fq}$, $\lambda_\textit{bc}$ and $\lambda_\textit{dc}$ are empirically set to 10, 10, 1 and 0.5, respectively. Other hyper-parameters are the same as the default setup in DN-DETR. 

\textbf{Training.} All models are trained on 4 GPUs with AdamW optimizer. We set the weight decay to $1 \times 10^{-4}$ and the initial learning rate to $1 \times 10^{-5}$. The total batch size is 16, with 8 images per domain.
The training has two steps. We first train the source-only model on the source domain dataset. This step takes 50 epochs, and we decrease the learning rate by a factor of 10 after 40 epochs. 
Then, the source-only model is fine-tuned on both source and target domain data.
For the Clipart$_{\texttt{all}}$ datasets, the training takes 30k iterations with a learning rate decay by a factor of 10 at the  20k iterations. For the other datasets, we train 20k iterations with the learning rate divided by 10 at the 10k iterations.

\begin{table}[t!]
\centering
\scriptsize
\renewcommand\arraystretch{1.1}
\setlength{\tabcolsep}{6.3pt}
\setlength{\aboverulesep}{0pt} 
\setlength{\belowrulesep}{0pt} 
\begin{tabular}{l|cccccc|c}
\toprule
\hspace{-0.5em}Method & bike & bird & car & cat & dog & person & mean \\
\Xhline{2\arrayrulewidth}
\hspace{-0.5em}\textit{\textcolor{gray}{WS Group}} & & & & & & & \\
PCL~\cite{8493315} & 6.7 & 28.8 & 20.2 & 9.5 & 5.4 & 27.4 & 16.3 \\
EDRN~\cite{DRN-WSOD_2020_ECCV} & 5.2 & 29.3 & 15.3 & 1.4 & 0.9 & 34.9 & 14.5 \\
\hline
\hspace{-0.5em}\textit{\textcolor{gray}{UDA Group}} & & & & & & & \\
SWDA~\cite{saito2019strong} & 82.3 & 55.9 & 46.5 & 32.7 & 35.5 & 66.7 & 53.3 \\
HTD~\cite{chen2020harmonizing} & 69.2 & 49.5 & 49.5 & 34.9 & 30.8 & 61.2 & 49.2 \\
ATF~\cite{he2020domain} & 78.8 & 59.9 & 47.9 & 41.0 & 34.8 & 66.9 & 54.9 \\
MCAR~\cite{zhao2020adaptive} & 87.9 & 52.1 & 51.8 & 41.6 & 33.8 & 68.8 & 56.0 \\
IIOD~\cite{9362301} & \textbf{95.8} & 54.3 & 48.3 & 42.4 & 35.1 & 65.8 & 56.9 \\
UMT~\cite{deng2021unbiased} & 88.2 & 55.3 & 51.7 & 39.8 & 43.6 & 69.9 & 58.1 \\
I$^3$Net~\cite{chen2021i3net} & 81.1 & 49.3 & 46.2 & 35.0 & 31.9 & 65.7 & 51.5 \\
VDD~\cite{wu2021vdd} & 90.0 & 56.6 & 49.2 & 39.5 & 38.8 & 65.3 & 56.6 \\
DBGL~\cite{chen2021dual} & 83.1 & 49.3 & 50.6 & 39.8 & 38.7 & 51.3 & 53.8 \\
AT~\cite{li2022cross} & 93.6 & 56.1 & \textbf{58.9} & 37.3 & 39.6 & 73.8 & 59.9 \\
\hline
\hspace{-0.5em}\textit{\textcolor{gray}{CDWS Group}} & & & & & & & \\
DT+PL~\cite{inoue2018cross} & 81.0 & 49.5 & 39.5 & 32.3 & 28.4 & 62.4 & 48.8 \\
ICCM~\cite{hou2021informative} & 86.6 & \textbf{64.2} & 52.6 & 32.4 & 41.2 & 67.4 & 57.4 \\
PLGE~\cite{ouyang2021pseudo} & 73.7 & 56.1 & 50.6 & 42.5 & 41.8 & \textbf{74.6} & 56.5 \\
H$^{2}$FA~\cite{xu2022h2fa} & 88.6 & 52.4 & 53.6 & 46.4 & \textbf{44.5} & 73.8 & 59.9 \\

\rowcolor[RGB]{220,220,220}DETR-GA & 80.5 & 59.5 & 55.4 & \textbf{48.9} & 43.4 & 72.0 & \textbf{60.0} \\
\bottomrule
\end{tabular}
\vspace{-1em}
\caption{Mean AP performance (\%) on Watercolor.}
\label{tab:watercolor}
\vspace{-1.5em}
\end{table}

\subsection{The effectiveness of DETR-GA}
\textbf{Comparison with the state-of-the-art methods.}
We compare the proposed DETR-GA with previous state-of-the-art methods. According to the training data and annotations, we classify all the methods into weakly supervised (``WSOD''), unsupervised domain adaptive (``UDAOD''), and CDWSOD methods, as shown in Table~\ref{tab:intro_methods}.

The results on Clipart$_{\texttt{all}}$, Clipart$_{\texttt{test}}$, Watercolor and Comic are summarized in Table~\ref{tab:clipart}, Table~\ref{tab:cliparttest}, Table~\ref{tab:comic} and Table~\ref{tab:watercolor}, respectively. We observe that on all these four datasets, the proposed DETR-GA achieves superior CDWSOD accuracy.  For example, on Clipart$_{\texttt{all}}$, DETR-GA achieves 79.4\% mAP, surpassing the strongest competitor (H$^2$FA) by +9.6\% mAP. In this paper, we report 79.4\%, 59.2\%, 60.0\% and 48.5\% mAP for CDWSOD (with VOC as the source domain) on Clipart$_{\texttt{all}}$, Clipart$_{\texttt{test}}$, Watercolor and Comic, respectively. These results outperform all the competing methods (sometimes by a large margin, \emph{e.g.}, on Clipart$_{\texttt{all}}$), setting new states of the art. 

\textbf{DETR-GA makes good utilization of the weak supervision.} Since the DETR-style detectors usually perform favorably against the pure-CNN detectors on the generic object detection tasks, a natural question is: whether the superiority of DETR-GA is because DETR itself
has higher base performance than the pure-CNN detectors. 

To figure out this question, we investigate the improvement over direct cross-domain deployment. In Table~\ref{tab:base}, the ``source-only'' row indicates the model is trained on the source domain and directly deployed on the target domain. For the pure-CNN detectors, we choose two methods, ``DT+PL''~\cite{inoue2018cross} and H$^{2}$FA~\cite{xu2022h2fa}. ``DT+PL is arguably the earliest and the most popular CDWSOD method, while H$^{2}$FA is the most recent state-of-the-art method. For fair comparison, we report their results based on the ResNet-101 backbone. From Table~\ref{tab:base}, we draw two observations as below:

First, under the direct cross-domain deployment, the achieved results of DETR-style detector and pure-CN detector are very close. On Clipart$_{\texttt{test}}$, Watercolor and Comic, DETR (DN-DETR, in particular) is slightly higher than Faster RCNN by 1.1\%, 0.8\% and 0.5\% mAP, respectively, while on Clipart$_{\texttt{all}}$, DETR is -0.2\% mAP lower. We thus infer that DETR does not always promise superiority on the cross-domain object detection task. 

Second, DETR-GA achieves larger improvement for CDWSOD. On these datasets, weakly-supervised domain adaptation brings 50.4\%, 28.6\%, 17.8\% and 28.1 mAP improvement for DETR, which are larger than (or comparable to) the improvement on the pure-CNN detector. Therefore, though the direct cross-domain deployment results are close, the proposed DETR-GA achieves higher CDWSOD accuracy than the pure-CNN detectors. It validates our key argument, \emph{i.e.}, DETR can better exploit image-level supervision for CDWSOD.

\begin{table}[t!]
\centering
\scriptsize
\setlength{\tabcolsep}{4.1pt}
\renewcommand\arraystretch{1.1}
\setlength{\aboverulesep}{0pt} 
\setlength{\belowrulesep}{0pt} 

\begin{tabular}{l|l|llll}
\Xhline{2\arrayrulewidth} 
Based & Method     & Clipart$_{\texttt{all}}$    & Clipart$_{\texttt{test}}$    & Watercolor    & Comic    \\
\Xhline{2\arrayrulewidth} 
\multirow{3}{*}{CNN}          & source-only & 29.2 & 29.5 & 41.4 & 19.9 \\
                              & DT+PL~\cite{inoue2018cross}       & 55.3 \tiny{\hlg{(+26.1)}} & 48.9 \tiny{\hlg{(+19.4)}} & 48.8 \tiny{\hlg{(+7.4)}} & 35.0 \tiny{\hlg{(+15.1)}}   \\
                              & H$^{2}$FA~\cite{xu2022h2fa}        & 69.8 \tiny{\hlg{(+40.6)}} & 55.3 \tiny{\hlg{(+25.8)}} & 59.9 \tiny{\hlg{(+18.5)}} & 46.4 \tiny{\hlg{(+26.5)}} \\
\Xhline{2\arrayrulewidth}                           
\multirow{3}{*}{DETR}         & source-only & 29.0 & 30.6 & 42.2 & 20.4 \\
                              & DETR-GA     & 79.4 \tiny{\hlg{(+50.4)}} & 59.2 \tiny{\hlg{(+28.6)}} & 60.0 \tiny{\hlg{(+17.8)}}  & 48.5 \tiny{\hlg{(+28.1)}} \\
                              & \textcolor{Gray}{oracle}      & \textcolor{Gray}{-----}  & \textcolor{Gray}{64.7} & \textcolor{Gray}{60.0}& \textcolor{Gray}{54.7}  \\
\bottomrule
\end{tabular}
\caption{While CNN and DETR detectors perform comparably for direct cross-domain deployment, DETR-GA brings larger improvement and achieves higher CDWSOD accuracy.}
\label{tab:base}
\vspace{-1.5em}
\end{table}


\subsection{Ablation study}\label{sec:ablation}
Table~\ref{tab:ablation} investigates the major components of DETR-GA through ablation on four benchmarks. We use CQ and FQ to indicate the class query and foreground query for brevity. The baseline adopts only common practices ~\cite{xu2022h2fa,wang2021exploring} for aligning the backbone features. We draw some important observations below. 

\textbf{DETR-GA gains individual benefit from each component}, particularly from the proposed CQ and FQ. 
Comparing Lines \textit{(c)}-\textit{(d)} against the source-only model in Line \textit{(a)}, we observe that each component alone brings noticeable improvement. Moreover, the CQ and FQ make the most prominent improvement. For example, on Clipart$_{\texttt{test}}$, using the CQ and FQ improve the accuracy by $+20.8\%$ and $+23.0\%$ mAP, respectively. 

\textbf{Combining these components achieves complementary benefits.}
For example, combining CQ and FQ (Line (f)) is higher than using only CQ (FQ) by + 5.2 (0.8) mAP. Based on ``CQ + FQ'', adding the common practices on the backbone further brings 0.9 and 4.8 mAP on Comic and Clipart$_\texttt{test}$, respectively.

\begin{table}[t!]
\centering
\scriptsize
\setlength{\tabcolsep}{9pt}
\renewcommand\arraystretch{1.1}
\setlength{\aboverulesep}{0pt} 
\setlength{\belowrulesep}{0pt}
\begin{tabular}{l|ccc|ll}
\toprule
  & baseline & CQ & FQ & Comic & Clipart$_{\texttt{test}}$ \\
\Xhline{2\arrayrulewidth}
\textit{(a)} &          &     &         & 20.4       & 30.6          \\
\textit{(b)} & \cmark       &     &     & 43.3 \tiny{\hlg{(+22.9)}}       & 49.3 \tiny{\hlg{(+18.7)}}     \\
\textit{(c)} &          & \cmark  &     & 42.5  \tiny{\hlg{(+22.1)}}     & 51.4 \tiny{\hlg{(+20.8)}}     \\
\textit{(d)} &          &     & \cmark  & 44.3  \tiny{\hlg{(+23.9)}}     & 53.6 \tiny{\hlg{(+23.0)}}     \\
\hline
\textit{(e)} & \cmark       & \cmark  &     & 47.7  \tiny{\hlg{(+27.3)}}   & 56.6 \tiny{\hlg{(+26.0)}}     \\
\textit{(f)} &          & \cmark  & \cmark  & 46.1  \tiny{\hlg{(+25.7)}}    & 54.4 \tiny{\hlg{(+23.8)}}     \\
\textit{(g)} & \cmark       &     & \cmark  & 47.8  \tiny{\hlg{(+27.4)}}    & 58.1 \tiny{\hlg{(+27.5)}}     \\
\hline
\textit{(h)} & \cmark       & \cmark  & \cmark  & 48.5 \tiny{\hlg{(+28.1)}}  & 59.2 \tiny{\hlg{(+28.6)}}    \\
\bottomrule
\end{tabular}
\vspace{-.3em}
\caption{Effectiveness of different modules in DETR-GA, where mean AP performance (\%) over all classes is reported. We use CQ and FQ to indicate the class query  and foreground query.}
\label{tab:ablation}
\end{table}

\begin{figure}[t]
\centering
\vspace{-.8em}
\includegraphics[width=0.9\linewidth]{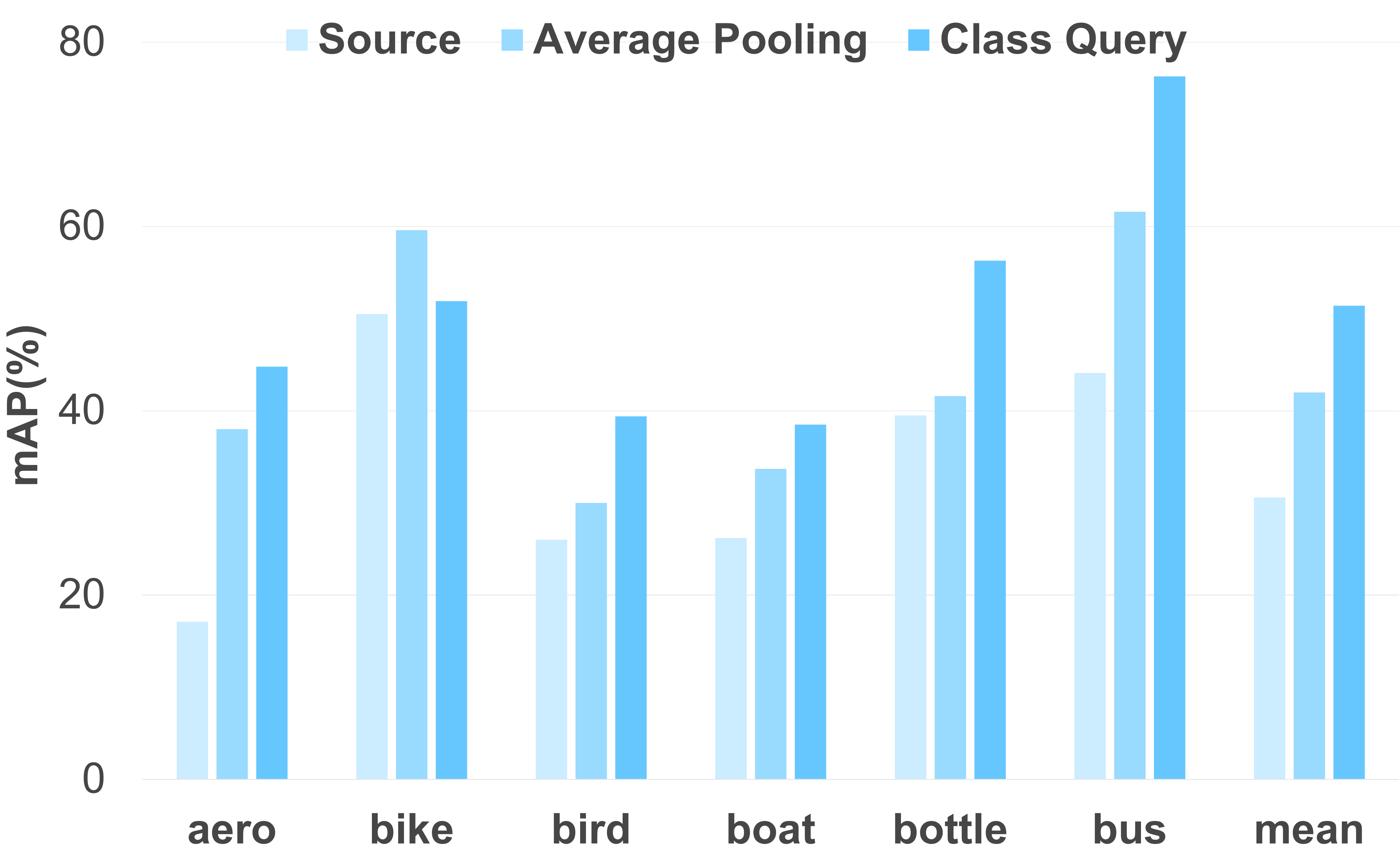}
\vspace{-.5em}
\caption{Mean AP (\%) performance on Clipart$_{\texttt{test}}$. We compare the class query with average pooling on the encoder. We only show the first 6 classes and average accuracy for brevity.}
\label{fig:abaltion1}
\vspace{-.5em}
\end{figure}

\textbf{Investigation on the class query.}
We compare two global aggregation methods for the encoder, \emph{i.e.}, the proposed class query and the popular average pooling. 
The results are shown in Figure~\ref{fig:abaltion1}.
It is observed that using average pooling achieves some improvement but is significantly inferior to the proposed class query. The reason is that the class query has some capability to exclude the distraction from non-relevant regions (as shown in Figure~\ref{fig:intro}) and thus facilitates better global aggregation.

\begin{table}[t!]
\centering
\vspace{-.3em}
\scriptsize
\renewcommand\arraystretch{1.1}
\setlength{\tabcolsep}{6.3pt}
\setlength{\aboverulesep}{0pt} 
\setlength{\belowrulesep}{0pt} 
\begin{tabular}{c|c|c|c}
\toprule
Foreground Query &  Position Embedding & Weight Sharing  & Map         \\
\Xhline{2\arrayrulewidth}
\cmark        &        &     & 51.5       \\
\cmark        & \cmark       &  \cmark   & 51.6      \\
\rowcolor[RGB]{220,220,220}\cmark        &        &     \cmark  & 53.6 \\
\bottomrule
\end{tabular}
\vspace{-.5em}
\caption{Effectiveness of different settings for foreground query. The mean AP performance (\%) on Clipart$_{\texttt{test}}$ is reported.}
\label{tab:samples}
\vspace{-2em}
\end{table}

\begin{figure}[t]
\centering
\vspace{-0.3em}
\includegraphics[width=\linewidth]{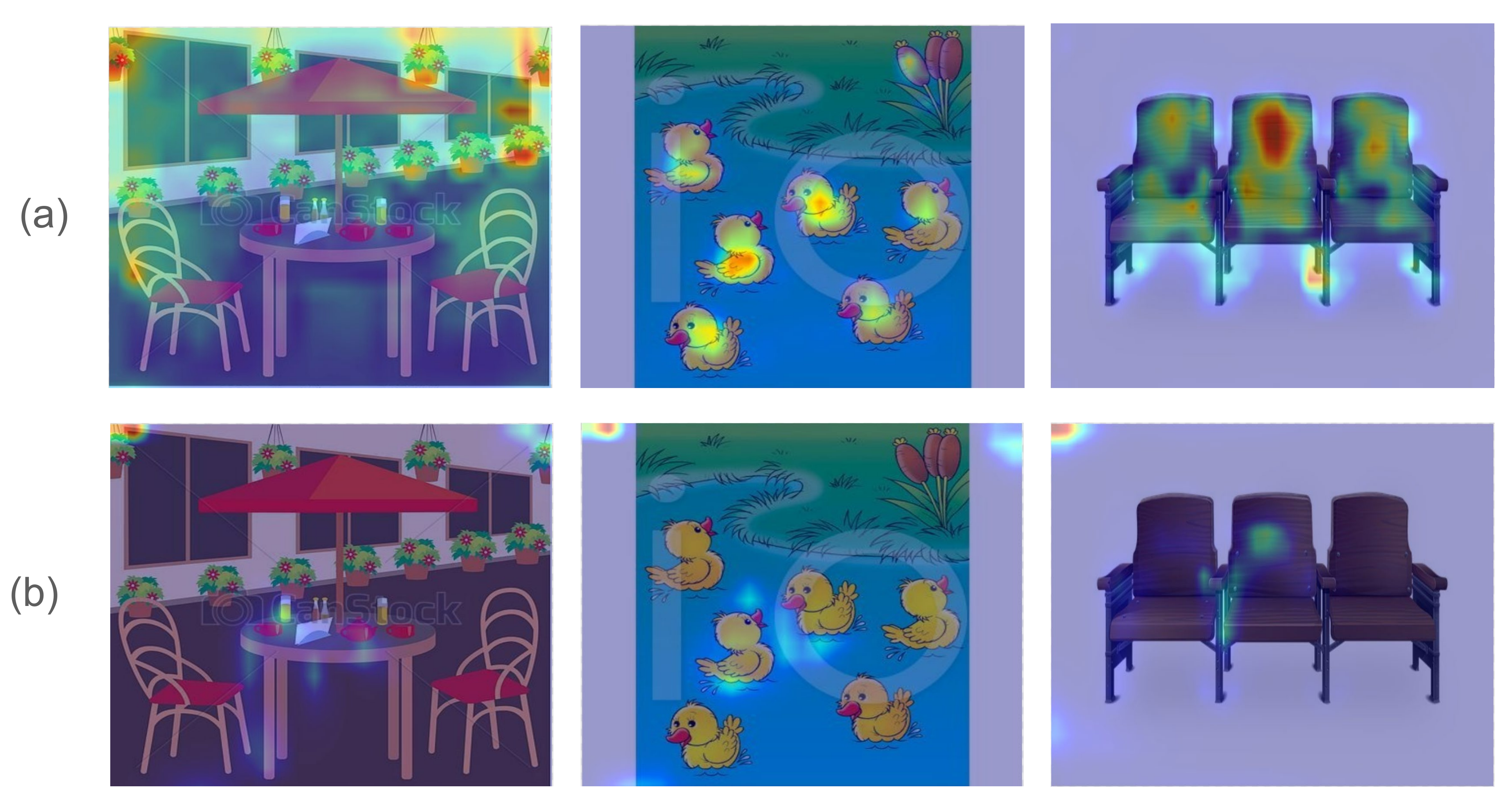}
\vspace{-1.5em}
\caption{The attention map of (a) the foreground query without position embedding, and (b) the foreground query with position embedding. Adding position embedding compromises the global aggregation capability of the foreground query.}
\label{fig:pos_attn}
\vspace{-1.5em}
\end{figure}

\textbf{Investigation on the foreground query.}
The foreground query is correlated with the object queries but has no position embedding. We validate these two designs in Table~\ref{tab:samples}. We observe that: 1) canceling the correlation between foreground query and object queries (the $1$st row) decreases the mAP by 2.1, and 2) using position embedding decreases the mAP by 2.0. 

Moreover, we visualize the attention map of the foreground query in Figure~\ref{fig:pos_attn}. From Figure~\ref{fig:pos_attn} (a), it is observed that the foreground query roughly attends to all the potential foreground. However, if we add position embedding to the foreground query (and still maintain the image-level supervision), the foreground query is prone to local focus and misses some important regions, as shown in Figure~\ref{fig:pos_attn} (b). 

\section{Conclusion}

In this work, we propose DETR with additional Global Aggregation (DETR-GA) for cross-domain weakly supervised object detection (CDWSOD). 
The main idea of DETR-GA is very straightforward: we add multiple class queries for the encoder and a foreground query for the decoder, respectively, to aggregate the semantics into image-level predictions.
The class queries in the encoder are able to aggregate the global semantics corresponding to the classes. 
And the foreground query in the decoder is correlated with the object queries, thus combining the strong and weak supervision spontaneously benefits domain alignment.
Extensive experiments show the superiority of DETR-GA over the competing state-of-the-art.

\textbf{Limitations.}
Deformable DETR is another popular DETR-style detector, which replaces the origin attention in DETR~\cite{carion2020end} with deformable attention. Although the current class query is compatible to the Deformable DETR encoder (in the supplementary material), we think exploring deformable-attention-based class query might fit it better.

\clearpage

{\small
\bibliographystyle{ieee_fullname}
\bibliography{camera_ready}
}

\end{document}